\pdfoutput=1

\documentclass[11pt]{article}

\usepackage{acl}
\usepackage{makecell}
\usepackage{times}
\usepackage{latexsym}
\usepackage{multirow}
\usepackage{hyperref}
\usepackage{booktabs}
\usepackage{graphicx}
\usepackage{soul} 
\usepackage{subfig}
\usepackage{epstopdf}
\usepackage{graphics} 
\usepackage{graphicx}
\usepackage{color}

\usepackage{hyperref}
\soulregister{\underline}7 
\soulregister{\cite}7 
\soulregister{\citep}7 
\soulregister{\citet}7 
\soulregister{\ref}7 
\soulregister{\pageref}7 

\usepackage{epsfig}
\usepackage{graphicx}
\usepackage{subfig} 
\usepackage[T1]{fontenc}
\usepackage{CJK}

\usepackage[utf8]{inputenc}

\usepackage{microtype}

\usepackage{inconsolata}
\usepackage{enumitem}

\title{Rich Semantic Knowledge Enhanced Large Language Models for Few-shot Chinese Spell Checking}

\author{  
	Ming Dong$^{1,2,3}$,  
	Yujing Chen$^{1,2,3}$,  
	Miao Zhang$^{4}$,  
	Hao Sun$^{1,2,3}$,  
	Tingting He$^{1,2,3,}$\thanks{~~Corresponding author.~~}$^{\ast}$  \\
	$^1$Hubei Provincial Key Laboratory of Artificial Intelligence and Smart Learning, \\
	$^2$National Language Resources Monitoring and Research Center for Network Media, \\
	$^3$School of Computer, Central China Normal University, Wuhan, China \\
	$^4$School of Computer Science and Information Engineering, Hubei University, Wuhan, China \\
	\texttt{\{dongming,haosun,tthe\}@ccnu.edu.cn}\\
	\texttt{\{yujingchen\}@mails.ccnu.edu.cn} \\
	\texttt{\{miaozhang\}@hubu.edu.cn}\\
}

\begin{document}
	\begin{CJK}{UTF8}{gbsn}
		
		\maketitle
		\begin{abstract}
			Chinese Spell Checking (CSC) is a widely used technology, which plays a vital role in speech to text (STT) and optical character recognition (OCR). Most of the existing CSC approaches relying on BERT architecture achieve excellent performance. However, limited by the scale of the foundation model, BERT-based method does not work well in few-shot scenarios, showing certain limitations in practical applications. In this paper, we explore using an in-context learning method named RS-LLM (\textbf{R}ich \textbf{S}emantic based LLMs) to introduce large language models (LLMs) as the foundation model. Besides, we study the impact of introducing various Chinese rich semantic information in our framework. We found that by introducing a small number of specific Chinese rich semantic structures, LLMs achieve better performance than most of the BERT-based model on few-shot CSC task. Furthermore, we conduct experiments on multiple datasets, and the experimental results verified the superiority of our proposed framework.
		\end{abstract}
		
		\section{Introduction}
		Spell checking (SC) aims to utilize intelligent methods to automatically identify and correct errors in text. This technology facilitates nature language processing applications to correct the errors from different text input systems, such as speech to text (STT) and optical character recognition (OCR). In recent years, SC has attracted tremendous attention from the research community \cite{chodorow2007detection, DBLP:conf/emnlp/MalmiKRMS19, DBLP:conf/emnlp/MallinsonSMG20}. Chinese spell checking (CSC) specifically refers to SC for Chinese text. Compared with the relatively complete SC technology, CSC still cannot be perfectly applied to various practical scenarios, and there are still many problems that need to be solved \cite{DBLP:conf/acl/ZhangZ0Q23}.
		
		As an ideogram, the usage and structure of Chinese are very different from English, which leads to different challenges for CSC and SC. First of all, the pronunciation of Chinese varies greatly, and it is difficult to easily infer the glyphs from the pronunciation. When listening to a piece of Chinese speech, if you do not understand the context, the text result obtained based on the speech is likely to contain a large number of homophones. In addition, the glyph structure of Chinese is more diverse, resulting in more types of errors. Therefore, CSC task mainly needs to handle two types of error texts. One is a text based on the STT system, which contains a large number of homophonic errors. The other is text generated by OCR-based systems that mainly contains glyph errors.  To address these two types of spelling errors, most of the existing studies use models based on BERT architecture, then introducing the external glyph-phonetic features \cite{DBLP:conf/emnlp/JiYQ21, DBLP:conf/acl/XuLZLWCHM21, DBLP:conf/emnlp/JiYQ21}.
		
		In practical Internet applications, the catchwords from hot topics vary rapidly and unlabeled incorrect sentences emerge constantly, resulting in few-shot scene for CSC. However, existing BERT-based models are difficult to be conduct in few-shot scene because of the limited scale of the foundation model. Large language models (LLMs) show remarkable ability on semantic analyzing, positioning them to become an optimal foundation model for CSC. This paper focus on CSC in few-shot scene. We build a Chinese rich semantic corpus (See details in Section \ref{sec:CRSCorpus}). Besides, we choose LLMs as foundation and integrate Chinese rich semantic knowledge by in-context learning. Furthermore, we conduct experiments on several datasets. The contributions of this study are summarized as follows:
		
		\begin{itemize}[itemsep=2pt,topsep=0pt,parsep=0pt]
			\item We propose an in-context learning based method to introduce LLMs to CSC task, which improves the performance of few-shot scenarios.
			\item We propose the paradigm of prompt template designing for CSC.
			\item We conduct experiment to compare different Chinese Rich Semantic structures. And we find the best structures for LLMs based CSC tasks.
		\end{itemize}
		
		\begin{figure*}[t]
			\centering
			\includegraphics[width=1.0\linewidth]{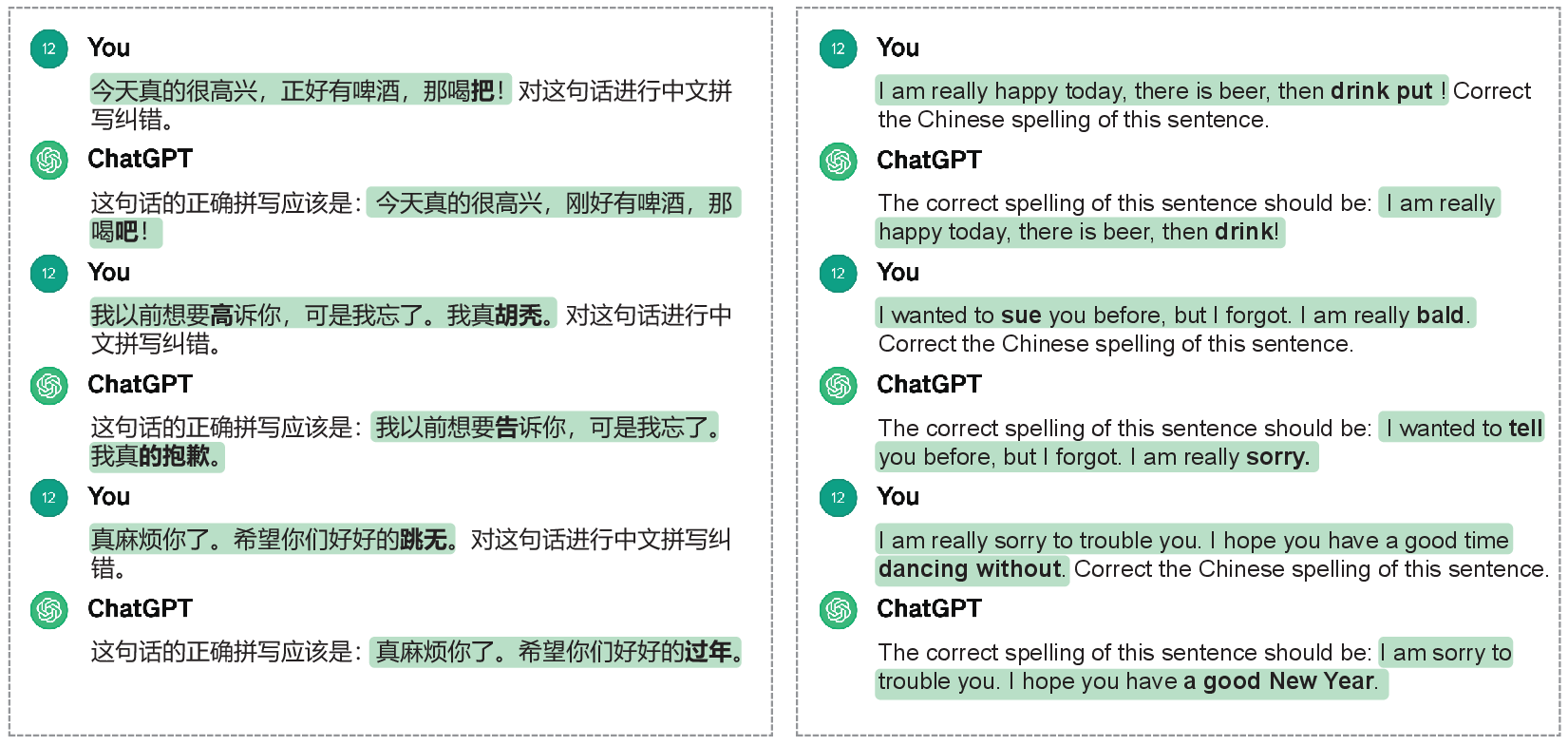}
			\caption{The performce of CSC tasks using LLM (ChatGPT) without specific restrictions on input prompts. }
			\label{fig:ChatGPT bad answer}
		\end{figure*}
		
		\section{Related Work}
		Due to the lack of parallel corpus training data, early CSC methods mainly rely on linguistic knowledge to manually design rule-based methods \cite{DBLP:conf/icml/ManguB97,jiang2012rule}. Subsequently, machine learning models become the main paradigm for CSC tasks \cite{DBLP:conf/acl-sighan/ChenLLWC13,DBLP:conf/acl-sighan/YuL14}. Machine learning typically employs language models, such as n-grams, to detect error locations. Then uses confusion sets and character similarities to correct potential misspelled characters and candidate correct characters, and finally scores replace sentences through the language model to determine the best correction solution \cite{DBLP:conf/acl-sighan/LiuCLDM13, DBLP:conf/acl-sighan/XieHZHHCH15}.
		
		The field of CSC advances significantly with the development of deep learning, particularly through pre-trained models like BERT \cite{DBLP:conf/naacl/DevlinCLT19}. Pre-trained models such as BERT are known for their context awareness and transfer learning capabilities. Most current CSC models with better performance use BERT as the baseline model. \citet{DBLP:conf/aclnut/HongYHLL19} innovates by modeling CSC as a BERT token classification task, utilizing the Confidence-Character Similarity Decoder (CSD). \citet{DBLP:conf/acl/ZhangHLL20} enhances this approach by combining error identification and correction losses with a soft mask strategy. Addressing the prevalent issue of phonetic and glyph similarities in spelling errors, the integration of these features with semantic information is now a primary research focus. \citet{DBLP:conf/acl/LiuYYZW20} suggests incorporating confusion sets into pre-training with a GRU network to better mimic real errors and model character sound-shape similarities. \citet{DBLP:conf/acl/XuLZLWCHM21} proposes a multi-modal approach to capture semantic, phonetic, and graphical information and the use of adaptive gating modules to merge semantic, phonetic, and glyph features in CSC. \citet{DBLP:conf/emnlp/JiYQ21} introduces SpellBERT, which integrates radical features into character representation using a graph convolution network. The SCOPE model \cite{DBLP:conf/emnlp/LiWMGY022} further develops this field by adding pronunciation prediction tasks in training, forging a deeper connection between semantics and phonetics, and employing iterative reasoning strategies to bolster CSC model performance.
		
		Recently, the development and advancement of LLMs have brought natural language processing to the next stage. \citet{DBLP:journals/corr/abs-2307-09007} analyzes the correction ability of the OpenAI$ \footnote{\url{https://openai.com/}}$'s existing LLMs and finds that they still fall short of the CSC capabilities of previous fine-tuned models.
		\section{Preliminary}
		\subsection{Chinese Rich Semantics}
		Chinese is the carrier of the inheritance and development of Chinese civilization. As a hieroglyphic script that has lasted for thousands of years, Chinese characters have rich semantic structures, including traditional characters, glyphs, phonetic, tones and other features. Chinese contains a large number of phonograms, whose pronunciation and meaning are contained in radicals. For example, "海 (sea)", "河 (river)" and "湖 (lake)" all have the radical "氵", related to water \cite{DBLP:conf/acl/SunLSMAHWL20}. For another example, "株 (zhū)", "诛 (zhū)" and "珠 (zhū)" all have the same phonetic tone "朱 (zhū)", so the pronunciation of these characters is similar. In essence, radicals can be regarded as a kind of classification label information. This classification information serves as traditional context. The additional supplement of semantic features has important semantic value. In addition to the features at the glyph level, Chinese also has features such as morpheme diversity, ambiguity, and structural diversity.
		
		\subsection{Task Definition}
		\textbf{Formulate definition of CSC:}
		CSC focuses on identifying and rectifying spelling errors in a given text sequence. Essentially, it involves processing a sequence of characters $X = \{x_1, x_2, ..., x_n\}$ and producing a corrected sequence $Y = \{y_1, y_2, ..., y_n\}$, where  $n$ is the number of word. Here,  $X $ denotes the initial text with potential errors, and  $Y$ signifies the amended, correct text. The two sequences  $X$ and  $Y $ have the same length. \\
		\textbf{Few-shot Learning on CSC:}
		Few-shot learning refers to provide a model named $L$ with $K$ pairs of contexts and corresponding required answers as examples, and then a context that requires model $L$ to reason. The goal of model $L$ is to generate an appropriate answer for the last context. During inference, model $L$ is guided by $K$ context-answer pairs without any updates to its parameters \cite{DBLP:conf/nips/BrownMRSKDNSSAA20}. 
		
		In few-shot learning for CSC task, we set $K$ to 3. Then we have a few-shot learning prompt $P$, which contains three sentences pairs:
		\begin{equation}
			P = \{p_{1}, p_{2}, p_{3}\}
		\end{equation}
		where \( p_{i} \) contains an incorrect sentence and its corresponding correct sentence.

		Then the sentence $X$ that need to be spell checked:
		\begin{equation}
			X = \{x_{1}, x_{2}, \ldots, x_{n}\}
		\end{equation}
		where \( x_{i} \) denotes a word of sentence $X$. $n$ is the length of sentence $X$. 
		
		For model $L$, the inputs are $P$ and $X$, and the output is the correct sentence $Y$ corresponding to the sentence $X$.
		\begin{equation}
			Y = L (P, X)
		\end{equation}
		\begin{equation}
			Y = \{y_{1}, y_{2}, \ldots, y_{n}\}
		\end{equation}
		where \( y_{i} \) denotes a word of sentence $Y$ and $n$ is the length of sentence $Y$. 
		\subsection{Difficulties of LLMs based CSC Tasks}
		\label{sec:dif}
		The versatility of LLMs gives them significant text polishing capabilities. Since there are no specific restrictions on input prompts, LLMs tend to perform freely in CSC tasks. However, free play of LLMs may result in LLMs outputting sentences that are completely grammatically correct. However, the sentences output by LLMs are different from the standards established by existing CSC datasets and evaluation indicators. Therefore, these existing traditional datasets are used without specific restrictions on input prompts. It becomes challenging to objectively and realistically evaluate the spell checking performance of LLMs. An example of using LLMs (ChatGPT) to perform a CSC task without setting specific restrictions on input prompts is shown in Fig.\ref{fig:ChatGPT bad answer}. From Fig.\ref{fig:ChatGPT bad answer}, it can be observed that two main problems are prone to occur in LLMs when performing CSC tasks, one is the length of the input sentence and the output sentence are inconsistent. For instance, the sentence "I am really bald (我真胡秃)", which should be corrected to "I am really confused (我真糊涂)". However, LLMs correct this sentence to "I am really sorry (我真的抱歉)". Another problem is that LLMs easily rewrite the input sentences semantically. For example, "I hope you have a good time dancing without (希望你们好好的跳无)", in which "dancing without (跳无)" should be corrected to "dancing (跳舞)". Instead, LLMs correct this sentence to "I hope you have a good New Year (希望你们好好的过年)".
		\begin{figure*}[t]
			\centering
			\includegraphics[width=1.0\linewidth]{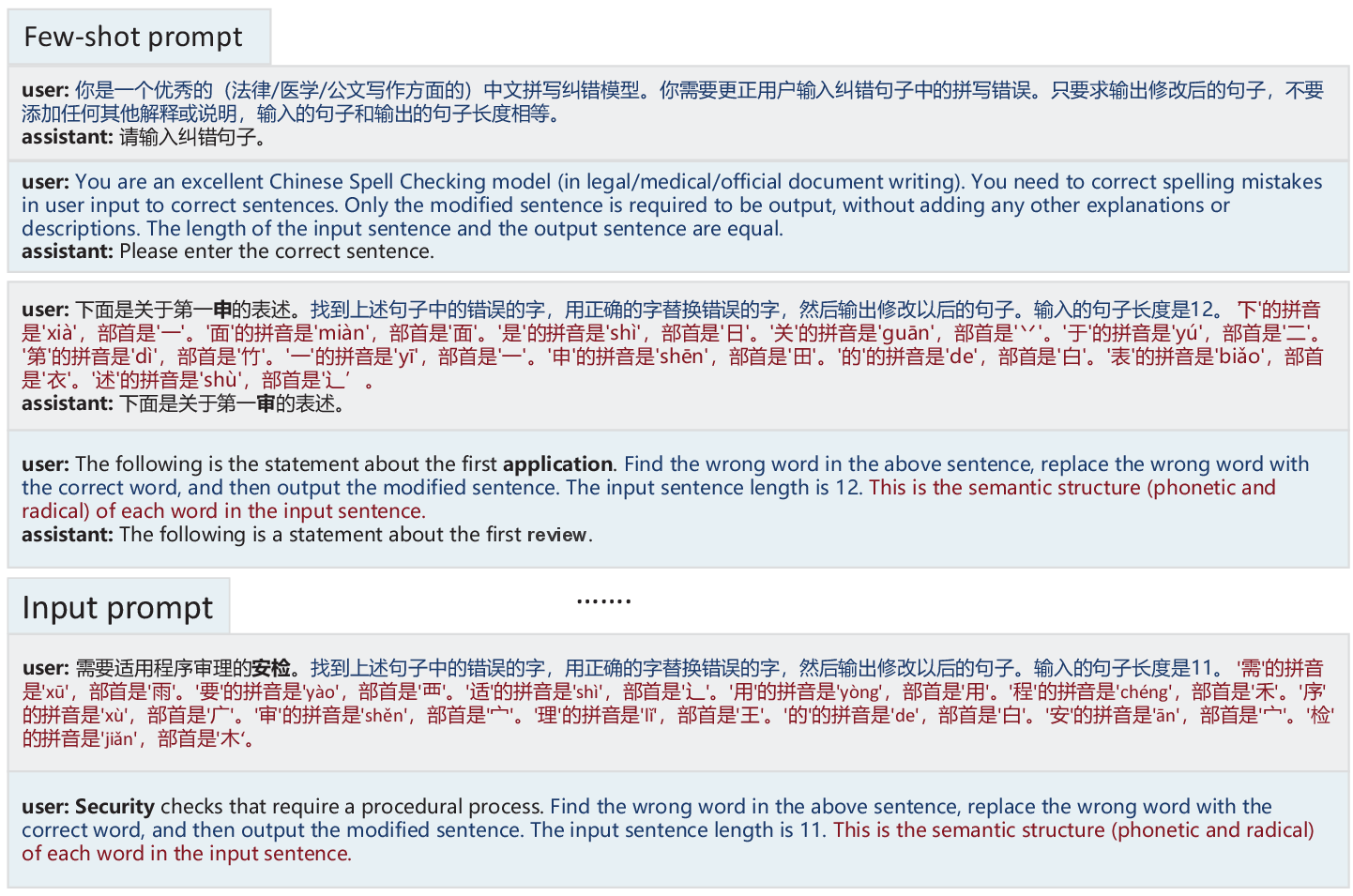}
			\caption{ Task-specific few-shot prompts for CSC tasks. We marked the \textcolor[RGB]{179,0,0}{semantic structure information (speech and radicals)} and \textcolor[RGB]{24,89,143}{key information related to the task features} in the prompts in different color. }
			\label{fig:prompt}
		\end{figure*}
		\section{Method}
		\subsection{Chinese Rich Semantic Corpus}
		\label{sec:CRSCorpus}
\begin{figure}[t]
			\centering
			\includegraphics[width=0.9\linewidth]{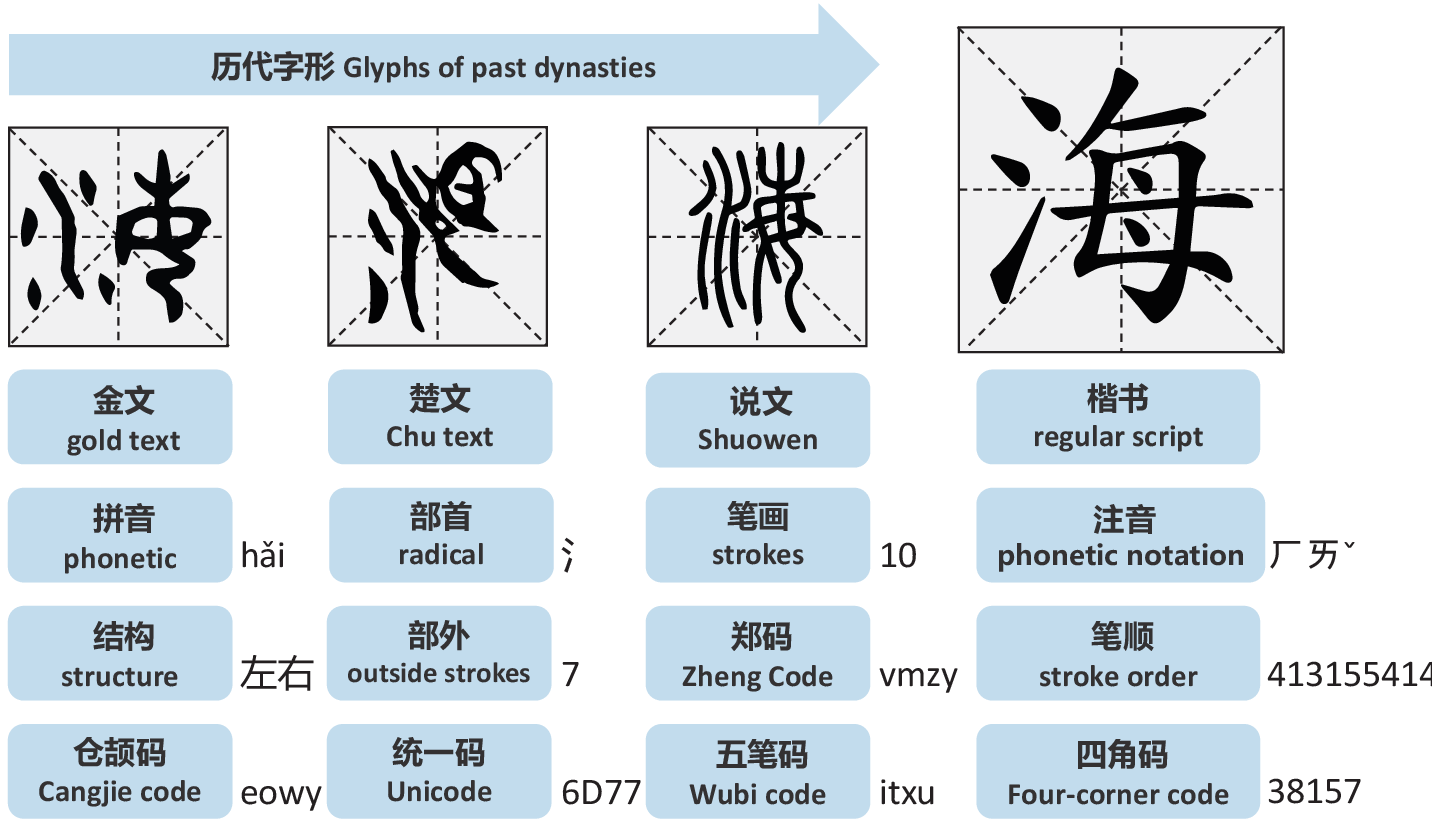}
			\caption{Various attributes of the word 海 (sea) in the Chinese Rich Semantics.}
			\label{fig:chinese-rich-semantics}
		\end{figure}

		Chinese, a logo-graphic language, inherently possesses a rich semantic depth in its character glyph, potentially enhancing the expressiveness of LLMs. Our work focuses on the GB2312 simplified Chinese coding table, a standard set by China's State Administration of Standards on May 1, 1981$ \footnote{\url{https://openstd.samr.gov.cn/bzgk/gb}}$. This table consists of 6,763 Chinese characters, divided into 3,755 primary and 3,008 secondary characters. It contains the most commonly used Chinese characters. In order to obtain the detailed information of this coding table, we collected various basic attributes of each Chinese word, such as its \textit{phonetic},\textit{ radical}, \textit{phonetic notation}, \textit{strokes}, \textit{outside strokes (the strokes except the radical)}, \textit{stroke order}, \textit{structure}, \textit{Unicode},\textit{ Wubi code}, \textit{Cangjie code}, \textit{Zheng code}, \textit{Four-corner code}, as well as \textit{glyph images from different historical periods}. Despite the large amount of data collected, we note issues with the quality and completeness of the data. To address these issues, we manually annotate the collected information to ensure a more accurate and comprehensive dataset. We give the attributes of the word 海 (sea) in the dataset as shown in Fig.\ref{fig:chinese-rich-semantics}. In order to understand and better utilize these properties, we classify these features into the following three categories:
		\begin{itemize}[itemsep=1pt,topsep=0pt,parsep=0pt]
			\item \textbf{Phonetic Features：}\textit{Phonetic} uses Latin letters to represent the pronunciation of Chinese characters. \textit{Phonetic notation} is a phonetic system that uses symbols to represent the speech of Chinese characters. \textit{Zheng Code} is a Chinese character input method that assigns codes based on the initials of the phonetic pronunciation.
			\item \textbf{Glyph Features：}\textit{Radical} is a category according to the type and side of the Chinese characters, and all the Chinese characters are bound to be classified in a certain radical. \textit{Strokes} refers to the number of lines needed to write Chinese character. \textit{Outside strokes} mean the number of lines needed to write Chinese character except the radical. \textit{Structure} refers to the internal organization of Chinese characters, including the arrangement of radicals, strokes, and components.
			\item \textbf{Input Coding Features：}\textit{Stroke Order} indicates the sequence in which strokes are written when forming a Chinese character. Proper stroke order is important for correct character writing. \textit{Cangjie code} assigns codes to characters based on their shapes and components. \textit{Unicode} is an international standard for character encoding that assigns unique codes to characters from various writing systems, including Chinese characters.\textit{ Wubi code} assigns codes based on the five basic strokes. \textit{Four-Corner code} is a Chinese character input method that assigns codes based on the shapes of the four corners of a character. 
			
		\end{itemize}
		We publish the information we collect at dropbox. $ \footnote{\url{https://www.dropbox.com/scl/fo/0r1jw4l1ex3w0lyojsfpf/AGVMdlOFpwDqoOFrxqWXJko?rlkey=18wke1vmhj6muwufvu3aazji2&st=vgh1drbi&dl=0}}$
		\subsection{In Context Learning for CSC}
		\subsubsection{Motivation}
		A variety of studies \cite{DBLP:conf/iclr/XieRL022,DBLP:conf/acl/DaiS0HMSW23,DBLP:conf/acl/BansalGDBKR23} have revealed that LLMs exhibit exceptional in-context learning capabilities \cite{DBLP:journals/corr/abs-2301-00234}.
		In-context learning can quickly improve the task-specific performance of LLMs by providing a limited set of task-related examples, which can quickly adapt to most LLMs without the need for separate training for each LLM. As mentioned in Section \ref{sec:dif}, when LLMs perform CSC tasks without specific task constraints on input prompts, the answers generated by LLMs are very likely to be inconsistent with the existing evaluation indicators of CSC tasks. Therefore, in order to accurately and objectively explore the performance ability of LLMs on CSC tasks, we design the task-specific prompts as shown in Fig.\ref{fig:prompt}. 
		
		\subsubsection{Prompt Design}
		First of all, we give the identity and task description in the prompt. According to \citet{li2023large}, we know that Role Attribution (``You are an excellent Chinese Spell Checking model") can effectively stimulate LLMs' comprehension. The task description requires the LLMs to only correct spelling mistakes, thus limiting the LLMs' semantic rephrasing of the input sentence. Since the evaluation of CSC task requires the input sentence and the output sentence to be same length, the LLMs are required not to add any other explanations and descriptions of output, so as to ensure the length remains unchanged.
		
		Then, we give three pairs of input sentences and their corrected sentences as examples, carefully selected from the corresponding training set of the dataset we use. From \citet{DBLP:conf/coling/LiuLCL10}, Chinese text errors are primarily caused by characters that are visually and phonetically similar. These three sentence pairs contain a set of phonetic errors, a set of glyph errors, and a set of correct sentences that do not need to be corrected.
		
		Next, we add specific task requirements to the input sentences of three pairs of sentences, which is different from the foremost task description. The foremost task description requires LLMs to correct the spelling errors of the input sentences. We concretize the task here, asking LLMs to find spelling errors in sentences and replace wrong word with correct one. This allows LLMs to have a clearer comprehension of CSC task. We append the length of the input sentence to the end to indicate the LLMs that we limit the length of the output sentence in the first prompt. We also include the phonetic and radical information of each Chinese character in the input sentence, drawn from Chinese Rich Semantic Corpus outlined in Section \ref{sec:CRSCorpus}. It is hoped that by adding this information to the prompt, the LLMs can strengthen their understanding of the input sentence. Therefore, LLMs can better correct phonetic and visual errors that may occur in the sentences. These are in the form of four pairs of historical dialogues that form a few-shot prompt input to LLMs, hoping to stimulate the in-context learning capabilities of LLMs at once.
		
		Considering the length limit of LLMs on the length of the input (including historical dialogue), and the deterioration of the semantic understanding of LLMs with long input lengths. For each sentence to be corrected in the test set, we clear the historical memory of the LLMs and add our few-shot prompt.
		
		\subsection{Introspection Mechanism}
		A significant challenge with using LLMs for spell checking is that LLMs tend to over-modify and arbitrarily change sentence lengths. In order to avoid the impact of this change on the evaluation indicators, we use a Introspection mechanism.
		
		Specifically, after LLMs generate the correction sentence, we send this answer and the original input sentence to LLMs again. LLMs are required to introspect two questions: 1) Whether the lengths of the two sentences are consistent. 2) Whether the added rich semantic information is effectively used in this error correction process. Only when the answers to both questions are "yes" will the answer be output as the final error correction result. Otherwise, the current conversation will be added to the historical conversation and provided as context to the model. In the conversation, it indicates that the sentence length in this answer does not match the input sentence length or the semantic information is not used, and then sends a reply request to LLMs again. In the introspection mechanism of this experiment, we set the maximum number of loops to 5. If the answer cannot be obtained after five requests to LLMs, it is judged that LLMs cannot correct the sentence and uses original input sentence without introspection as the answer. And consider that the model's ability to understand long context will deteriorate. We only add the latest round of dialogue in each loop within the context of the original design.
		
		\section{Experiments}
		
		\subsection{Datasets and Evaluation Metrics}
		\textbf{Datasets:} To evaluate the effectiveness of our proposed method, we choose two widely used datasets. The first, SIGHAN15 \cite{DBLP:conf/acl-sighan/TsengLCC15}, consists of handwritten samples from learners of Chinese as a Second Language (CSL). These samples provide a rich source of real-world language usage by non-native speakers, offering insights into common errors and patterns in learning Chinese. The second dataset \cite{DBLP:journals/talip/LvCGAYF23} is specialized and segmented into three distinct domains to provide a more comprehensive understanding of language use in specific contexts. For the LAW domain, data is sourced from the stems and choices of multiple-choice questions in judicial examinations, reflecting the formal and technical language of the legal field. The MED domain encompasses data from question-and-answer pairs drawn from online medical consultations, showcasing the specific terminology and communication style in healthcare. The ODW (Official Document Writing) domain includes data from various official documents such as news, policies, and state reports on national conditions, representing formal and structured writing styles. The statistics of the test data from these four datasets we used are shown in Table \ref{table:data}.\\
		\begin{table}[h!]
			\centering
			\begin{tabular}{c|ccc}
				\toprule
				\textbf{Test Data} & \textbf{\#Sent} & \textbf{Avg. Length} & \textbf{\#Errors} \\
				\midrule
				SIGHAN15 & 1100 & 30.6 & 703 \\
				LAW & 500 & 29.7 & 356 \\
				MED & 500 & 49.6 & 345 \\
				ODW & 500 & 40.5 & 403 \\
				\bottomrule
			\end{tabular}
			\caption{Statistical information regarding the dataset in our experiments, which includes the total count of sentences (\#Sent), the average length of these sentences (Avg. Length), and the total number of spelling mistakes (\#Errors).}
			\label{table:data}
		\end{table}
		
		\textbf{Evaluation Metrics:} By following the existing work \cite{DBLP:conf/acl/XuLZLWCHM21,DBLP:journals/talip/LvCGAYF23}, we evaluate the performance on two metrics: detection and correction. For detection, a sentence is considered correct if it successfully identifies all spelling errors. For correction, the model not only identifies but also rectifies all erroneous characters by replacing them with the correct ones. We provide accuracy, precision, recall, and F1-scores for both metrics.
		
		\subsection{Baselines}
		We choose two widely used LLMs as foundational models.\\
		\textbf{gpt-3.5-turbo}$ \footnote{\url{https://platform.openai.com/docs/api}}$: One of a series of LLMs provided by OpenAI, which can be accessed through the API. The gpt-3.5-turbo serves as the underlying module for ChatGPT and is trained on GPT-3 using Reinforcement Learning from Human Feedback (RLHF).\\
		\textbf{ChatGLM2-6B}$ \footnote{\url{https://huggingface.co/THUDM/chatglm2-6b}}$: An open-source bilingual (Chinese-English) chat model. ChatGLM2-6B employs a GLM-based \cite{du2022glm} hybrid objective function and has been pre-trained on 1.4 trillion bilingual tokens and human preference alignment training.\\
		We choose several advanced CSC models as baselines:\\
		\textbf{BERT} \cite{DBLP:conf/naacl/DevlinCLT19}: BERT encodes the input sentence to get semantic information, followed by using a classifier to select the correct character from the vocabulary.\\
		\textbf{ChineseBERT} \cite{DBLP:conf/acl/SunLSMAHWL20}: ChineseBERT encodes the input sentence to get semantic, phonetic, and graphical information, then use a classifier to select the correct character from the vocabulary.\\
		\textbf{ReaLiSe} \cite{DBLP:conf/acl/XuLZLWCHM21}: This CSC model captures semantic, phonetic, and graphical information of input characters using multimodality for prediction.\\
		\textbf{Scope} \cite{DBLP:conf/emnlp/LiWMGY022}: This CSC model introduces an auxiliary task of Chinese pronunciation prediction (CPP) to improve CSC task.\\
		\subsection{Implementation Details}
		For all methods, the settings of hyper-parameters follow the optimal parameters in the open source code corresponding to the model. In order to achieve few-shot scenarios, all contrast experiments randomly selected 10 samples of data from the training set, trained for 1000 epochs, and then selected the best performance. The LLMs experiment uses the same few-shot prompt designed for CSC task as RS-LLMs, but does not add semantic information. All experiments were conducted on a RTX-4090 with 24G memory.
		
		\begin{table}[t]
			\centering
			\begin{tabular}{c|ccc}
				\toprule
				\textbf{Data}  & \textbf{without prompt} & \textbf{with prompt} \\
				\midrule
				SIGHAN15 & 578 & 724 \\
				LAW & 83 & 125 \\
				MED  & 237 & 262 \\
				ODW  & 294 & 347 \\
				\bottomrule
			\end{tabular}
			\caption{The number of sentences whose length does not change using specific prompts and not using specific prompts on different data sets. Experimental result statistics are based on gpt-3.5-turbo.}
			\label{table:len}
		\end{table}
		
		\begin{table*}[h!]
			\small
			\begin{center}
				\begin{tabular}{c|c|cccc|cccc}
					\toprule
					\multirow{2}{*}{\textbf{Dataset}}& 
					\multirow{2}{*}{\textbf{Method}} & 
					\multicolumn{4}{c|}{\textbf{Detection Level}} &  
					\multicolumn{4}{c}{\textbf{Correction Level}}\\
					& & \textbf{Acc.} & \textbf{Pre.} &\textbf{ Rec.} & \textbf{F1} & \textbf{Acc.} & \textbf{Pre.} &\textbf{ Rec.} & \textbf{F1} \\
					\noalign{\smallskip} 
					\hline
					\noalign{\smallskip} 
					\multirow{8}{*}{SIGHAN15}
					& BERT &17.8  & 15.1 & 16.1 & 15.8  & 15.4 & 3.6 & 2.4 & 2.8 \\
					& ChineseBERT & 16.2 & 10.4 & 12.9 & 11.5 & 13.2& 2.7 & 5.5 & 3.6 \\
					& ReaLiSe & 27.5 &16.2  & 18.6 & 17.3  &22.3  &3.6  & 3.1 & 3.4\\
					& Scope &\underline{64.7} & \underline{61.7} & \underline{34.0} & \underline{43.9 }&\underline{58.1} & \underline{37.2} &\underline{ 20.5} & \underline{26.5} \\
					\noalign{\smallskip} 
					\cline{2-10}
					\noalign{\smallskip} 
					& ChatGLM2-6B & 18.3 & 7.1 & 10.1 &8.3$\pm$1.1 & 17.0 & 3.3 & 5.3 & 4.1$\pm$2.8\\
					& RS-ChatGLM2-6B & 30.7 & 14.9 & 18.0 &16.3$\pm$2.3 & 28.6 & 12.4 & 15.6 & 13.9$\pm$3.9 \\
					& gpt-3.5-turbo & 36.4 & 22.3 & 33.0 & 26.6$\pm$3.1 & 34.3 & 19.3 & 28.6 & 23.1$\pm$3.5 \\
					& RS-gpt-3.5-turbo & \textbf{50.6} & \textbf{32.5} & \textbf{41.6} & \textbf{36.5$\pm$4.7} &\textbf{ 48.1}& \textbf{31.2} & \textbf{40.8} & \textbf{35.4$\pm$5.9}  \\
					\noalign{\smallskip} 
					\hline
					\noalign{\smallskip} 
					\multirow{8}{*}{LAW}
					& BERT &14.4 & 8.8 & 12.3 & 10.3 & 12.7 & 1.35 & 0.4 & 0.6\\
					& ChineseBERT & 15.0 & 12.3 & 14.3 & 13.2 & 13.5 & 0.8 & 1.6 & 1.1 \\
					& ReaLiSe &23.3 & 15.4 & 18.2 & 16.7 & 20.2 & 1.6 & 1.5 &1.5 \\
					& Scope &\underline{66.2} &\underline{50.4} & \underline{48.6} & \underline{49.5} & \underline{58.6} & \underline{35.0} & \underline{33.7} & \underline{34.3} \\
					\noalign{\smallskip} 
					\cline{2-10}
					\noalign{\smallskip} 
					& ChatGLM2-6B & 36.6 & 18.5 & 25.5 & 21.4$\pm$2.7 & 34.8 & 15.9 & 21.9 & 18.5$\pm$3.3 \\  
					& RS-ChatGLM2-6B & 45.2 & 24.2 & 25.1 & 24.6$\pm$3.4 & 40.4 & 22.7 & 24.8& 23.7$\pm$3.2 \\
					& gpt-3.5-turbo & 48.8 & 30.2 & 38.8 & 34.0$\pm$3.6 & 46.2 & 26.2 & 33.7 & 29.5$\pm$3.6 \\
					& RS-gpt-3.5-turbo & \textbf{64.6} & \textbf{46.6} & \textbf{56.1} & \textbf{50.9$\pm$9.4} & \textbf{60.8} & \textbf{40.4} & \textbf{48.6} & \textbf{44.1$\pm$8.7} \\
					\hline
					\noalign{\smallskip} 
					\multirow{8}{*}{MED}
					& BERT  & 14.6 & 7.1 & 13.3 & 9.2 & 12.6 & 1.8 & 0.4 & 0.6\\
					& ChineseBERT  &14.0 &8.1 &10.4  &9.1 & 11.2 &0.6 & 1.5&0.9\\
					& ReaLiSe  &26.2 & 9.7 & 16.8 & 12.3 & 18.6 & 1.5 & 0.9 &1.1\\
					& Scope &\underline{66.4} &\underline{45.6} & \underline{53.9} & \underline{49.4} &\underline{56.5} &\underline{27.3}& \underline{32.3} &\underline{ 29.6 }\\
					\noalign{\smallskip} 
					\cline{2-10}
					\noalign{\smallskip} 
					& ChatGLM2-6B & 31.2 & 15.7 & 23.0 & 18.7$\pm$5.6 & 28.4 & 14.7 & 19.2 & 16.7$\pm$6.5 \\
					& RS-ChatGLM2-6B & 45.2 & 30.4 & 33.3 & 31.8$\pm$6.7 & 39.9 & 22.6 & 30.6 & 26.0$\pm$4.2 \\
					& gpt-3.5-turbo  & 41.4 & 16.8 & 30.1 & 21.5$\pm$1.2 & 38.6 & 13.3 & 23.9 & 17.1$\pm$2.7 \\
					& RS-gpt-3.5-turbo  &\textbf{56.0} &\textbf{31.0 }& \textbf{45.2} &\textbf{36.8$\pm$8.7} &\textbf{ 43.3}& \textbf{25.3 }&\textbf{ 39.3 }& \textbf{30.8$\pm$6.8} \\
					\noalign{\smallskip} 
					\hline
					\noalign{\smallskip} 
					\multirow{8}{*}{ODW}
					& BERT   & 16.8 & 13.2 & 15.6 & 14.3 & 13.1 &4.4 & 1.5 & 2.3\\
					& ChineseBERT  &15.9 & 12.2& 14.2 &13.1 & 11.6 & 2.0& 3.8&2.7\\
					& ReaLiSe  &30.2 & 18.5 & 26.8 & 21.9 & 25.4 & 5.4 & 4.2 &4.7\\
					& Scope &\underline{75.0} & \underline{65.5} &\underline{ 58.7} & \underline{62.0} &\underline{70.8} &\underline{ 56.5} &\underline{ 50.7} & \underline{53.5}  \\
					\noalign{\smallskip} 
					\cline{2-10}
					\noalign{\smallskip} 
					& ChatGLM2-6B & 47.6 & 28.6 & 32.5 & 30.4$\pm$6.5 & 44.2 & 27.0 & 31.1 & 28.9$\pm$7.8 \\
					& RS-ChatGLM2-6B & 56.4 & 32.1 & 37.8 & 34.7$\pm$2.7 & 50.8 & 29.8 & 32.4 & 31.0$\pm$3.2 \\
					& gpt-3.5-turbo & 63.0 & 45.8 & 50.0 & 47.8$\pm$8.5 & 59.2 & 39.2 & 42.8 & 40.9$\pm$7.9  \\ 
					& RS-gpt-3.5-turbo & \textbf{72.4} & \textbf{59.1} & \textbf{64.9} & \textbf{61.9$\pm$2.6} &\textbf{ 70.2} &\textbf{ 50.2} & \textbf{57.6} & \textbf{53.6$\pm$5.1} \\
					\bottomrule
				\end{tabular}
				\caption{The performance of all baselines and RS-LLMs. RS-gpt-3.5-turbo means RS-LLM on gpt-3.5-turbo and RS-ChatGLM2-6B means RS-LLM on ChatGLM2-6B. ChatGLM2-6B and gpt-3.5-turbo only utilize identical few-shot prompts as RS-LLM, without semantic prompt and introspection.\textbf{The bold information} indicates the best results except  \underline{Scope} (Scope is the best BERT-based model in few-shot scene), and $\pm$ indicates the error band of the results. }
				\label{table:result}
			\end{center}
		\end{table*}
		
		\begin{figure*}[t]
			\centering
			\includegraphics[width=0.4\linewidth]{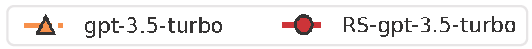}
			\setcounter{subfigure}{0}
			\subfloat{
				\includegraphics[scale=0.3]{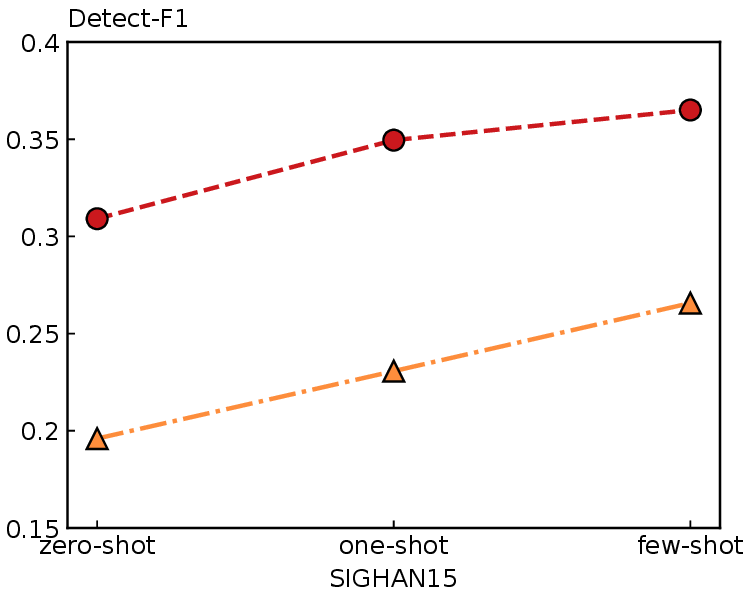}}
			\subfloat{
				\includegraphics[scale=0.3]{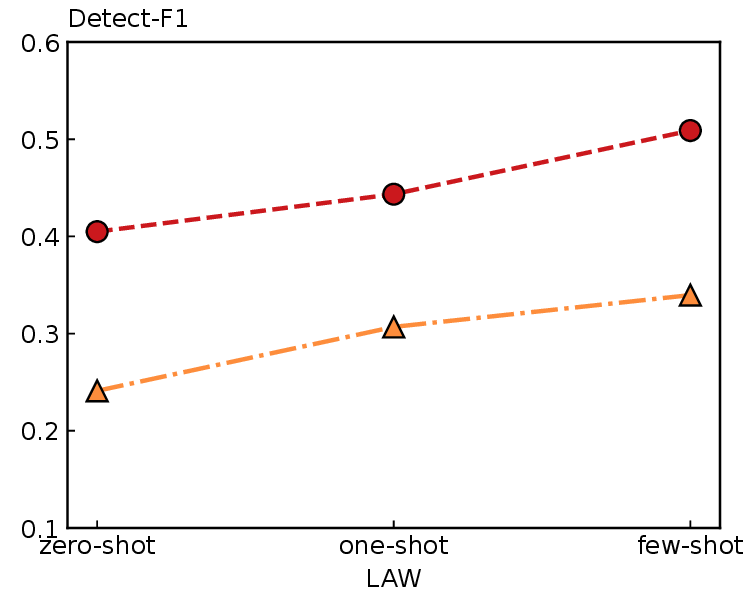}}
			\subfloat{
				\includegraphics[scale=0.3]{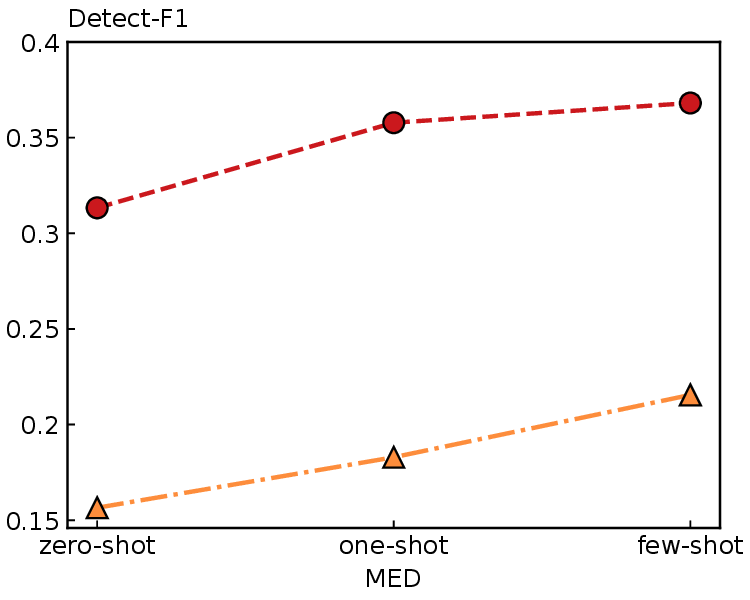}}
			\subfloat{
				\includegraphics[scale=0.3]{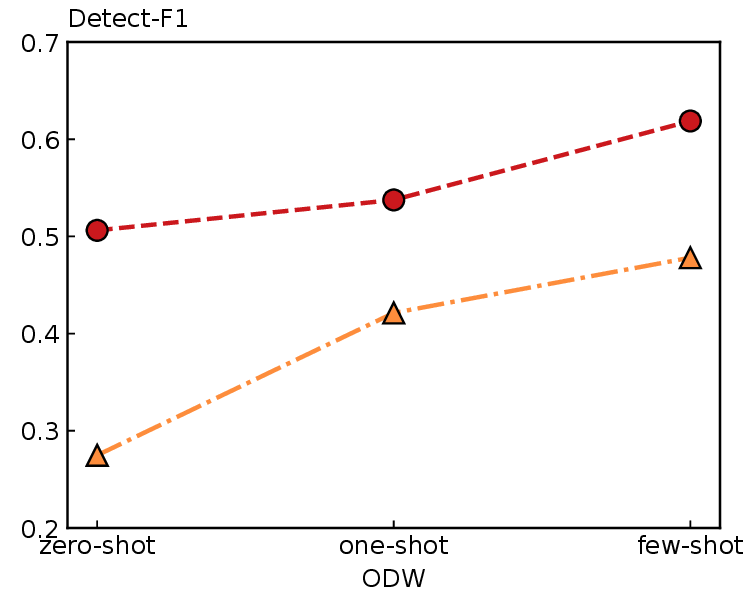}}
			\caption{Detect-F1 score trends on the test dataset. }
			\label{fig:detect} 
		\end{figure*}
		
		\begin{figure*}[t]
			\centering
			\subfloat{
				\includegraphics[scale=0.3]{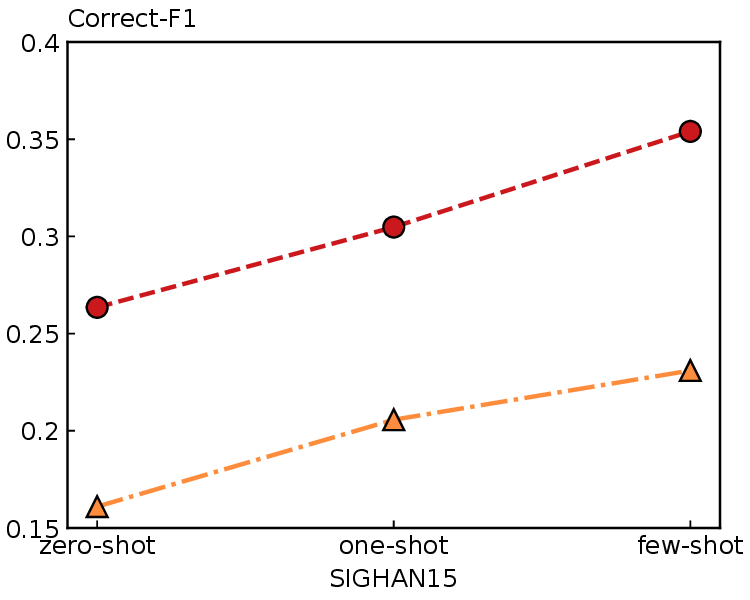}}
			\subfloat{
				\includegraphics[scale=0.3]{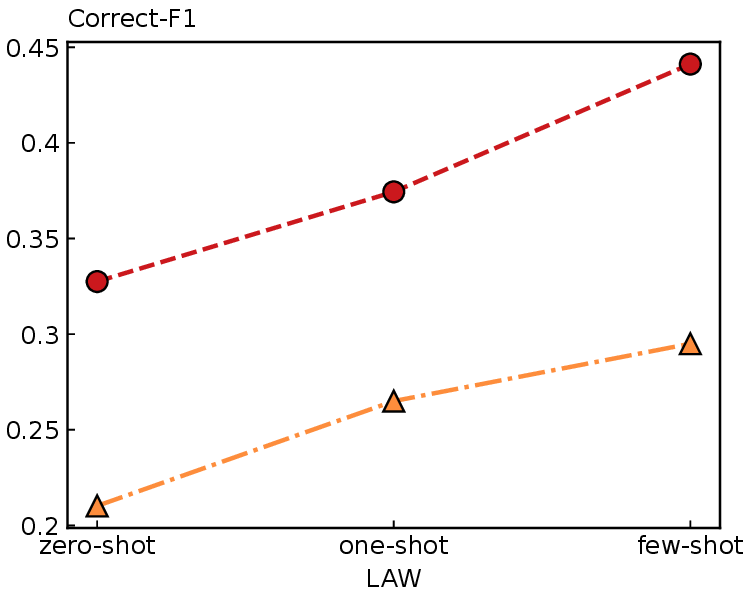}}
			\subfloat{
				\includegraphics[scale=0.3]{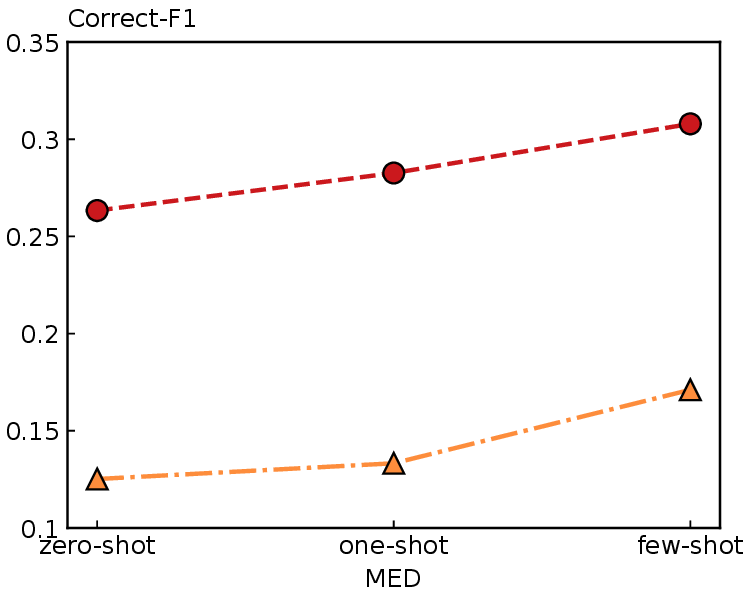}}
			\subfloat{
				\includegraphics[scale=0.3]{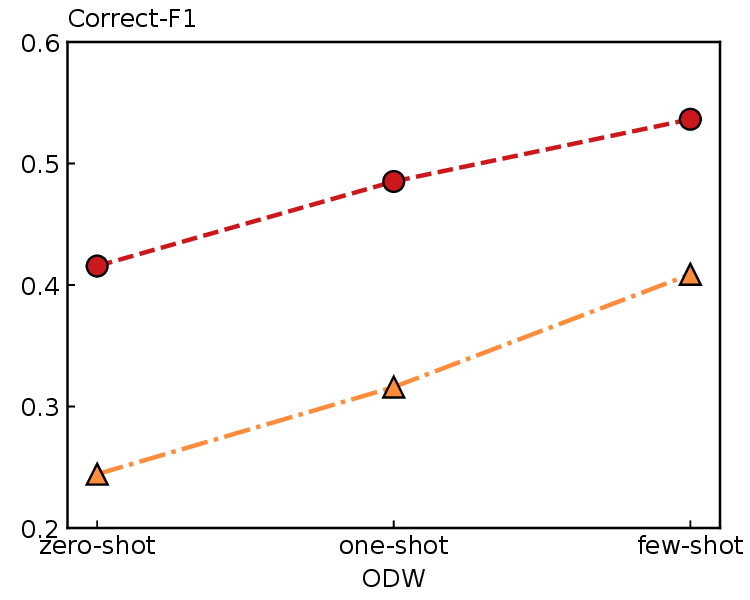}}
			\caption{The impact of different number of prompt samples. }
			\label{fig:correct} 
		\end{figure*}
		\subsection{Experimental Results}
		Table \ref{table:result} shows the evaluation results of our RS-LLMs comparing to other models on the test sets of SIGHAN15, LAW, MED, and ODW. It is observed that our RS-LLM consistently outperforms all baselines on all datasets in all metrics. Especially compared with LLM, LLM uses the same few-shot prompt as RS-LLM but does not add semantic information when performing CSC tasks. The experimental results of this period verify the effectiveness and superiority of our semantic information on LLM for CSC task.
		
		From time to time experiments, we found that the experimental results are not completely consistent because the performance of online APIs varies. Therefore, we use $\pm$ in the results to explain the error band of the experiment.
		
		As shown in Table \ref{table:result}, our spell-checking approach, RS-LLM on ChatGLM2-6B, achieves a notable 8.0\% improvement in error detection and an 9.8\% increase in correction on SIGHAN 15 using Rich Semantic, compared to the standard LLM. Impressively, RS-LLM on gpt-3.5-turbo registers a 9.9\% boost in F1-score for detection and a 12.3\% leap in correction. In the LAW dataset, RS-LLM on ChatGLM2-6B beats ChatGLM2-6B by 5.8\% in detection and 6.6\% in correction, while RS-LLM on gpt-3.5-turbo outperforms gpt-3.5-turbo by 16.9\% and 14.6\%, respectively. The trend continues in the MED dataset, where RS-LLM on ChatGLM2-6B surpasses ChatGLM2-6B by 13.1\% in detection and 9.3\% in correction, and RS-LLM on gpt-3.5-turbo exceeds gpt-3.5-turbo by 15.3\% and 13.7\%. On the ODW dataset, our method also shows significant gains, with RS-LLM on ChatGLM2-6B leading ChatGLM2-6B by 6.9\% in detection and 5.6\% in correction, and RS-LLM on gpt-3.5-turbo outdoing gpt-3.5-turbo by 14.1\% and 12.7\%.
		
		\subsection{Analyses and Discussions}
		
		In order to verify the effectiveness of the prompt sentence, we count the number of sentences whose length remained the same with and without a specific prompt, and the experimental results are shown in Table \ref{table:len}. It is obvious that LLM performs better than most of the BERT-based CSC models at both the detection and correction levels. The ability of the BERT-based CSC model to detect and correct erroneous characters is overly dependent on the training data of CSC, especially the ability to correct erroneous characters. The LLMs based approach shows better generalization in the few-shot scenario. However, we find that LLM's performance on the CSC task heavily depends on the foundation model. When the scale of parameters of the foundation model is larger, the model performs better on the CSC task. Although the performance of RS-LLM cannot currently outperform Scope, we believe that with the continuous update of the base model performance, the RS-LLM method will continue to improve and show a better improvement trend.
		
		\subsubsection{Impact of Different In-context Learning Approaches}
		Fig.\ref{fig:detect} and Fig.\ref{fig:correct} show the evaluation results of different in-context learning approaches. From Fig.\ref{fig:detect} and Fig.\ref{fig:correct}, we discover that the performance of CSC task on LLMs improves as the number of examples increasing, both in terms of detection and correction. Additionally, it's clear that RS-LLM is more effective than most BERT-based models in terms of zero-shot, one-shot, and few-shot scene. Such experimental results reflect the effectiveness of our in-context learning strategy designed for CSC task. The specific experimental results are shown in Table \ref{table:shot} in Appendix \ref{app:table}.
		
		\subsubsection{Impact of Different Chinese Rich Semantic Information}
		In order to further study the impact of semantic information on LLMs when performing CSC tasks.  We design to use phonetic information, radical information, structural information and strokes information. Each of these four types of information is added to our zero-shot prompt, one-shot prompt, and few-shot prompt. To explore the impact of these four prompts on the test set of SIGHAN15, LAW, MED, and ODW. The specific experimental results are Table \ref{table:few} in Appendix \ref{app:table}.
		
		The experimental results show that individual phonetic, radical, structural, and stroke information leads to improvements in CSC task across all datasets. Notably, the phonetic and radical information contribute the most significant enhancements, followed by structural information. While strokes information does show some improvement, it's relatively puny compared to the others.
		\section{Conclusion and Future Work}
		In this paper, we introduce an in-context learning method named RS-LLM for CSC task, one of whose core components is using Chinese Rich Semantics in LLMs for CSC task. RS-LLM utilizes adding a small number of specific Chinese rich semantic information into a specific few-shot prompt set for CSC task, aiming to allow LLMs to have a fuller understanding of the semantics when doing CSC task. Experimental results show that this LLMs-independent approach can help existing LLMs better recognize and correct phonetically and visually erroneous characters in CSC tasks. Considering the impact of the similarity of errors in the few-shot prompt and errors in the sentence on LLMs' understanding of the sentence when LLMs perform CSC task. In the future, we will try to construct dynamic prompt for each sample through semantic similarity retrieval.
		\section*{Limitations}
		
		To verify the effectiveness of RS-LLM, we conducted extensive experiments on two benchmark datasets of different domains and scales. The results indicate that RS-LLM delivers SOTA results in few-shot scenarios. Since most of the errors in the CSC dataset are attributed to visual and phonetical errors, we incorporate phonetic and radical information into the prompt template. However, it is difficult to ensure that the existing manually formulated prompt templates are optimal, and the optimal prompt sentences for CSC require further research. Furthermore, relevant examples have been carefully selected to enable LLMs to identify potential visual and speech errors in a small number of scenarios. No consideration was given to the ability to motivate the LLMs through semantic similarity. We recognize these two limitations and plan to address them in future research efforts.
		
		\section*{Ethics Statement}
		We adhere to and advocate for the principles outlined in the ACL Code of Ethics. The primary focus of this paper is on the task of CSC, with an objective to enhance the performance of LLMs in this task by incorporating semantic knowledge into the template. The datasets utilized in our research are obtained from openly published sources, ensuring they are free from privacy or ethical concerns. In our approach, we consciously avoid introducing or magnifying any social or ethical biases in the model or data. Consequently, we anticipate no direct social or ethical challenges as a result of our research.
		\section*{Acknowledgments}
		This work was partly supported by China Postdoctoral Science Foundation (No. 2023M731253) and Hubei Provincial Natural Science Foundation (No. 2023AFB487).
		
		\bibliography{ref}
		
		\appendix
		
		\section{The Detailed Experimental Results}
		
		\begin{table*}[h!]
			\small
			\begin{center}
				\begin{tabular}{c|c|cccc|cccc}
					\toprule
					\multirow{2}{*}{\textbf{Dataset}}& 
					\multirow{2}{*}{\textbf{Method}} & 
					\multicolumn{4}{c|}{\textbf{Detection Level}} &  
					\multicolumn{4}{c}{\textbf{Correction Level}}\\
					
					& & \textbf{Acc.} & \textbf{Pre.} &\textbf{ Rec.} & \textbf{F1} & \textbf{Acc.} & \textbf{Pre.} &\textbf{ Rec.} & \textbf{F1} \\
					
					\hline
					\multirow{6}{*}{SIGHAN15}
					& gpt-3.5-turbo & 36.4 & 22.3 & 33.0 & 26.6 & 34.3 & 19.3 & 28.6 & 23.1 \\
					& gpt-3.5-turbo+拼音& 45.7 & 31.2 & \textbf{46.7} & \textbf{37.4} & 43.0 & 27.5 & \textbf{41.2} & 33.0 \\
					& gpt-3.5-turbo+部首 & 48.1 & 29.7 & 36.8 & 32.8 & 45.6 & 26.6 & 31.8 & 28.4 \\
					& gpt-3.5-turbo+结构  & 45.2 & 27.8 & 39.4 & 32.6 & 43.6 & 25.4 & 36.0 & 29.8 \\
					& gpt-3.5-turbo+笔画 & 38.7 & 23.3 & 35.3 & 28.1 & 36.5 & 20.4 & 30.9 & 24.5 \\
					& RS-gpt-3.5-turbo & \textbf{50.6} & \textbf{32.5} & 41.6 & 36.5 & \textbf{48.1} & \textbf{31.2} & 40.8 & \textbf{35.4}  \\
					
					\hline
					
					\multirow{6}{*}{LAW}
					& gpt-3.5-turbo & 48.8 & 30.2 & 38.8 & 34.0 & 46.2 & 26.2 & 33.7 & 29.5 \\
					& gpt-3.5-turbo+拼音 & 64.0 & 45.4 & 48.6 & 47.0 & 60.8 & 39.6 & 42.4 & 40.9 \\
					& gpt-3.5-turbo+部首 & \textbf{65.0} & 45.7 & 49.4 & 47.5 & \textbf{61.4} & 39.1 & 42.4 & 40.7 \\
					& gpt-3.5-turbo+结构 & 62.4 & 41.8 & 43.1 & 42.5 & 59.6 & 36.5 & 37.7 & 37.1 \\
					& gpt-3.5-turbo+笔画 & 60.2 & 40.9 & 47.8 & 44.1 & 55.4 & 32.9 & 38.4 & 35.4 \\
					& RS-gpt-3.5-turbo & 64.6 & \textbf{46.6} & \textbf{56.1} & \textbf{50.9} & 60.8 & \textbf{40.4} & \textbf{48.6} &\textbf{ 44.1} \\
					\hline
					\multirow{6}{*}{MED} 
					& gpt-3.5-turbo  & 41.4 & 16.8 & 30.1 & 21.5 & 38.6 & 13.3 & 23.9 & 17.1 \\
					& gpt-3.5-turbo+拼音  & 52.6 & \textbf{31.0 }& 37.9 & 34.1 & \textbf{48.2} & 20.9 & 32.5 & 25.5 \\
					& gpt-3.5-turbo+部首  & 50.0 & 30.3 & 37.3 & 33.4 & 46.2 & 19.1 & 29.3 & 23.1 \\
					& gpt-3.5-turbo+结构  & 46.6 & 26.0 & 31.5 & 28.5 & 44.1 & 16.6 & 26.9 & 20.5 \\
					& gpt-3.5-turbo+笔画 & 42.4 & 24.9 & 30.5 & 27.5 & 40.0 & 16.5 & 28.1 & 20.8 \\
					& RS-gpt-3.5-turbo  & \textbf{56.0} & \textbf{31.0} &\textbf{ 45.2 }&\textbf{ 36.8} & 43.3 &\textbf{ 25.3} & \textbf{39.3 }& \textbf{30.8} \\
					\hline
					\multirow{6}{*}{ODW} 
					& gpt-3.5-turbo & 63.0 & 45.8 & 50.0 & 47.8 & 59.2 & 39.2 & 42.8 & 40.9  \\ 
					& gpt-3.5-turbo+拼音 & 70.4 & 55.7 & 58.0 & 56.8 & 67.2 & 49.8 & 51.9 & 50.8 \\
					& gpt-3.5-turbo+部首 & 70.8 & 55.0 & 60.3 & 57.6 & 66.0 & 46.7 & 51.2 & 48.8 \\
					& gpt-3.5-turbo+结构 & 70.8 & 55.2 & 58.4 & 56.8 & 66.2 & 46.9 & 49.6 & 48.2 \\
					& gpt-3.5-turbo+笔画 & 69.4 & 53.4 & 54.6 & 54.0 & 65.8 & 46.6 & 47.7 & 47.2 \\
					& RS-gpt-3.5-turbo & \textbf{72.4} & \textbf{59.1} &\textbf{ 64.9} & \textbf{61.9} & \textbf{70.2} & \textbf{50.2} & \textbf{57.6} & \textbf{53.6} \\
					\bottomrule
				\end{tabular}
				\caption{The detailed impact of different rich semantic structures on few-shot learning. RS-gpt-3.5-turbo means RS-LLM on gpt-3.5-turbo. '拼音' means phonetic information, '部首' means radical information, '结构' means structural information, and '笔画' means strokes information. }
				\label{table:few}
			\end{center}
		\end{table*}
		
		\begin{table*}[h!]
			\small
			\begin{center}
				\begin{tabular}{c|c|c|cccc|cccc}
					\toprule
					\multirow{2}{*}{\shortstack{\textbf{In-context Learning} \\ \textbf{Approaches}}}&
					\multirow{2}{*}{\textbf{Dataset}}& 
					\multirow{2}{*}{\textbf{Method}} & 
					\multicolumn{4}{c|}{\textbf{Detection Level}} &  
					\multicolumn{4}{c}{\textbf{Correction Level}}\\
					
					&&& \textbf{Acc.} & \textbf{Pre.} &\textbf{ Rec.} & \textbf{F1} & \textbf{Acc.} & \textbf{Pre.} &\textbf{ Rec.} & \textbf{F1} \\
					\hline
					\multirow{8}{*}{zero-shot}
					&\multirow{2}{*}{SIGHAN15}
					
					& gpt-3.5-turbo & 25.9 & 15.6 & 26.1 & 19.6 & 23.9 & 13.2 & 22.0 & 16.1 \\
					&& RS-gpt-3.5-turbo & \textbf{38.6} &\textbf{ 25.4 }& \textbf{40.1} & \textbf{30.9 }& \textbf{35.6} & \textbf{21.4} & \textbf{34.2} & \textbf{26.4} \\
					\cline{2-11}
					
					&\multirow{2}{*}{LAW}
					& gpt-3.5-turbo & 32.2 & 19.9 & 30.6 & 24.1 & 30.2 & 17.4 & 26.7 & 21.0 \\
					&& RS-gpt-3.5-turbo & \textbf{54.8} & \textbf{36.7} & \textbf{45.1} & \textbf{40.5} & \textbf{50.4} & \textbf{29.7} & \textbf{36.5} & \textbf{32.8} \\
					\cline{2-11}
					&\multirow{2}{*}{MED} 
					& gpt-3.5-turbo & 24.6 & 12.1 & 22.1 & 15.7 & 22.6 & 9.7 & 17.7 & 12.5 \\
					&& RS-gpt-3.5-turbo & 43.0 & \textbf{25.1} & \textbf{41.6} & \textbf{31.3} &\textbf{ 40.0} & \textbf{21.1} & \textbf{35.0} &\textbf{ 26.3} \\		
					
					\cline{2-11}
					&\multirow{2}{*}{ODW} 
					& gpt-3.5-turbo & 33.8 & 22.9 & 34.4 & 27.5 & 31.8 & 20.4 & 30.5 & 24.4 \\
					&& RS-gpt-3.5-turbo & \textbf{60.6} & \textbf{50.0} & \textbf{51.3} & \textbf{50.6} & \textbf{55.7 }& \textbf{41.1} & \textbf{42.1} &\textbf{ 41.6} \\
					
					\hline
					\multirow{8}{*}{one-shot}
					&\multirow{2}{*}{SIGHAN15}
					& gpt-3.5-turbo & 34.5 & 19.2 & 29.0 & 23.1 & 32.9 & 17.1 & 25.9 & 20.6 \\
					&& RS-gpt-3.5-turbo & \textbf{46.4} & \textbf{30.0} & \textbf{42.0} & \textbf{35.0} &\textbf{ 43.7} & \textbf{26.1} & \textbf{36.6} & \textbf{30.5} \\
					\cline{2-11}
					
					&\multirow{2}{*}{LAW}
					& gpt-3.5-turbo & 41.2 & 26.1 & 37.3 & 30.7 & 38.6 & 22.5 & 32.2 & 26.5 \\
					&& RS-gpt-3.5-turbo & \textbf{59.8} & \textbf{38.1} & \textbf{52.9} & \textbf{44.3} &\textbf{ 56.8} & \textbf{32.2 }& \textbf{44.7} &\textbf{ 37.4} \\
					\cline{2-11}
					&\multirow{2}{*}{MED} 
					& gpt-3.5-turbo & 36.2 & 14.1 & 26.1 & 18.3 & 23.0 & 10.3 & 19.0 & 13.3 \\
					&& RS-gpt-3.5-turbo & \textbf{51.4} & \textbf{30.0} & \textbf{44.3} & \textbf{35.8} & \textbf{47.2} &\textbf{ 23.7} & \textbf{35.0} & \textbf{28.3} \\			
					
					\cline{2-11}
					&\multirow{2}{*}{ODW} 
					& gpt-3.5-turbo & 44.9 & 38.8 & 46.0 & 42.1 & 42.1 & 28.2 & 36.0 & 31.6 \\
					&& RS-gpt-3.5-turbo & \textbf{64.8} & \textbf{49.5} & \textbf{58.8} & \textbf{53.8} & \textbf{61.8} &\textbf{ 44.7} & 53.1 &\textbf{ 48.5} \\

					\hline
					\multirow{8}{*}{few-shot}
					&\multirow{2}{*}{SIGHAN15}
					
					& gpt-3.5-turbo & 36.4 & 22.3 & 33.0 & 26.6 & 34.3 & 19.3 & 28.6 & 23.1 \\
					&& RS-gpt-3.5-turbo & \textbf{50.6} & \textbf{32.5} & 41.6 & 36.5 & \textbf{48.1} & \textbf{31.2} & 40.8 & \textbf{35.4}  \\
					\cline{2-11}
					
					&\multirow{2}{*}{LAW}
					& gpt-3.5-turbo & 48.8 & 30.2 & 38.8 & 34.0 & 46.2 & 26.2 & 33.7 & 29.5 \\
					&& RS-gpt-3.5-turbo & \textbf{64.6} & \textbf{46.6} & \textbf{56.1} & \textbf{50.9} &\textbf{ 60.8 }& \textbf{40.4} & \textbf{48.6} &\textbf{ 44.1} \\ 
					\cline{2-11}
					&\multirow{2}{*}{MED} 
					& gpt-3.5-turbo  & 41.4 & 16.8 & 30.1 & 21.5 & 38.6 & 13.3 & 23.9 & 17.1 \\
					&& RS-gpt-3.5-turbo  & \textbf{56.0} & \textbf{31.0} &\textbf{ 45.2 }&\textbf{ 36.8} & \textbf{43.3} &\textbf{ 25.3} & \textbf{39.3 }& \textbf{30.8} \\		
					
					\cline{2-11}
					&\multirow{2}{*}{ODW} 
					& gpt-3.5-turbo & 63.0 & 45.8 & 50.0 & 47.8 & 59.2 & 39.2 & 42.8 & 40.9  \\ 
					&& RS-gpt-3.5-turbo & \textbf{72.4} & \textbf{59.1} &\textbf{ 64.9} & \textbf{61.9} & \textbf{70.2} & \textbf{50.2} & \textbf{57.6} & \textbf{53.6} \\
					\bottomrule
				\end{tabular}
				\caption{The detailed performance of different number of prompt samples. RS-gpt-3.5-turbo means RS-LLM on gpt-3.5-turbo. }
				\label{table:shot}
			\end{center}
		\end{table*}

		In this section, we introduce the specific experimental results of different in-context learning methods under different semantic information.
		\label{app:table}
	\end{CJK}
\end{document}